\documentclass{article} % For LaTeX2e
\usepackage{iclr2016_conference,times}
\usepackage{graphicx}
\usepackage{subfig}
\usepackage{amsmath}
\usepackage{floatrow}
\usepackage{amsfonts}
\usepackage{multirow}
\usepackage{float}
\usepackage{algorithm}
\usepackage{algpseudocode}
\usepackage[innercaption]{sidecap}
\usepackage{pifont}
\usepackage{epstopdf}
\usepackage{tikz}
\sidecaptionvpos{figure}{t}

\title{Adding Gradient Noise Improves Learning\\ for Very Deep Networks}

\author{
Arvind Neelakantan\thanks{First two authors contributed equally. Work was done when all authors were at Google, Inc.}, Luke Vilnis\footnotemark[1]\\
College of Information and Computer Sciences\\
University of Massachusetts Amherst\\
\texttt{\{arvind,luke\}@cs.umass.edu} \\
\AND
Quoc V. Le, Ilya Sutskever, Lukasz Kaiser, Karol Kurach \\
{Google Brain} \\
\texttt{\{qvl,ilyasu,lukaszkaiser,kkurach\}@google.com}
\AND
James Martens \\
{University of Toronto} \\
\texttt{jmartens@cs.toronto.edu}
}

\begin{document}
\maketitle
\begin{abstract}
% Deep Networks have achieved impressive results in many speech
% recognition, computer vision and language processing
% applications. Despite this success, training deep networks is still
% difficult, and deep models tend to overfit training data. In this
% paper, we study the effectiveness of adding noise to the gradient
% during training, especially with stochastic gradient descent. On a
% variety of tasks, ranging from sequence-to-sequence models to memory
% networks, adding noise to gradient is an effective mechanism to
% improve the stability of training and circumvent overfitting.
Deep feedforward and recurrent networks have achieved impressive
results in many perception and language processing applications.  This
success is partially attributed to architectural innovations such as
convolutional and long short-term memory networks. The main motivation for
these architectural innovations is that they capture better domain
knowledge, and importantly are easier to optimize than more basic
architectures. Recently, more complex architectures such as Neural
Turing Machines and Memory Networks have been proposed for tasks
including question answering and general computation, creating a new
set of optimization challenges.  In this paper, we discuss a
low-overhead and easy-to-implement technique of adding gradient noise
which we find to be surprisingly effective when training these
very deep architectures. The technique not only helps to avoid
overfitting, but also can result in lower training loss.
This method alone allows a fully-connected 20-layer deep
network to be trained with standard gradient descent, even starting
from a poor initialization.  We see consistent improvements for many
complex models, including a 72\% relative reduction in error rate over a
carefully-tuned baseline on a challenging question-answering task, and
a doubling of the number of accurate binary multiplication models learned
across 7,000 random restarts. We encourage
further application of this technique to additional complex modern
architectures.
\end{abstract}

\section{Introduction}

Deep neural networks have shown remarkable success in diverse domains
including image recognition \citep{KrizhevskySH12}, speech recognition
\citep{38131} and language processing applications
\citep{SutskeverVL14,BahdanauCB14}. This broad success comes from a
confluence of several factors. First, the creation of massive labeled
datasets has allowed deep networks to demonstrate their advantages in expressiveness
and scalability. The increase in computing power has
also enabled training of far larger networks with more
forgiving optimization dynamics
\citep{DBLP:conf/aistats/ChoromanskaHMAL15}. Additionally,
architectures such as convolutional networks \citep{lecun1998gradient}
and long short-term memory networks \citep{Hochreiter:1997} have
proven to be easier to optimize than classical feedforward and
recurrent models. Finally, the success of deep networks is also a
result of the development of \emph{simple} and \emph{broadly applicable} learning techniques such as
dropout \citep{Srivastava:2014:DSW:2627435.2670313}, ReLUs
\citep{DBLP:conf/icml/NairH10}, gradient clipping \citep{pascanu2013difficulty,Graves13},
optimization and weight initialization strategies
\citep{glorot2010understanding,sutskever2013importance,DBLP:journals/corr/HeZR015}. 
%The fact that these techniques are simple is important to  their success because it is easy for practitioners to experiment with them. Developing a simple method that can be %generally applicable and simple is the main goal of this paper. 

% these models dont work on lot og problems -> new models proposed -> they are hard to optimize even with all the tricks -> we propose a simple trick that seems to benefit a large class of models

%Despite the success of standard deep networks in a large set of
%problem domains, this success has not always translated to more
%complex tasks, such as question answering and algorithmic tasks like
%sorting, arithmetic, and program induction. 
Recent work has aimed to push neural network learning into more
challenging domains, such as question answering or program
induction. These more complicated problems demand more complicated architectures (e.g.,
\cite{graves2014neural,sukhbaatar2015weakly}) thereby posing new
optimization challenges. In order to achieve good performance,
researchers have reported the necessity of additional techniques such as 
supervision in intermediate steps 
\citep{weston2014memory}, warmstarts~\citep{DBLP:journals/corr/PengLLW15}, random restarts, and the removal of certain
activation functions in early stages of training
\citep{sukhbaatar2015weakly}.

% For example, while training end-to-end memory networks which are neural networks augmented with an external memory the authors find it hard to optimize their model and use multiple random restarts to achieve good results.

A recurring theme in recent works is that commonly-used optimization
techniques are not always sufficient to robustly optimize the
models. In this work, we explore a simple technique of adding annealed
Gaussian noise to the gradient, which we find to be surprisingly
effective in training deep neural networks with stochastic gradient
descent. While there is a long tradition of adding random weight noise
in classical neural networks, it has been under-explored
in the optimization of modern deep architectures. In contrast to
theoretical and empirical results on the regularizing effects of
conventional stochastic gradient descent, we find that in practice the
added noise can actually help us achieve lower training loss by
encouraging active exploration of parameter space. This exploration proves
especially necessary and fruitful when optimizing neural network
models containing many layers or complex latent structures.

% While simple and easily-optimized models such as convolutional networks and LSTMs have 

% While this has enabled  have been broadly 

% This success has not been translated to more 

% cite learning to execute?

% To the best of our knowledge, our added noise schedule has not 
% been used before in the training of deep networks. 

The main contribution of this work is to demonstrate the broad
applicability of this simple method to the training of many complex
modern neural architectures. Furthermore, to the best of our knowledge, our added noise schedule has not 
been used before in the training of deep networks. 
% While our particular noise schedule has not, to the best of our
% knowledge, been used before in the training of deep networks, the
% contribution of this work is mainly to demonstrate the broad
% applicability of this simple method to the training of many complex
% modern neural architectures. 
We consistently see improvement from
injected gradient noise when optimizing a wide variety of models,
including very deep fully-connected networks, and special-purpose
architectures for question answering and algorithm learning. For
example, this method allows us to escape a poor initialization and successfully 
train a 20-layer rectifier network
on MNIST with standard gradient descent.
It also enables a 72\% relative
reduction in error in question-answering, and
doubles the number of accurate binary multiplication models learned
across 7,000 random restarts. We hope that practitioners will see similar improvements in
their own research by adding this simple technique, implementable in a
single line of code, to their repertoire.

\section{Related work}

Adding random noise to the weights, gradient, or the hidden units has
been a known technique amongst neural network practitioners for many
years (e.g.,~\cite{an1996effects}). However, the use of gradient noise
has been rare and its benefits have not been fully documented with
modern deep networks.

% While modern deep learning practitioners have learned to train classical feedforward and recurrent architectures without resorting to such tricks, 

Weight noise \citep{Steijvers96arecurrent} and adaptive weight noise \citep{graves2011practical,blundell2015weight}, which usually maintains a Gaussian variational posterior over network weights, similarly aim to improve learning by added noise during training. 
They normally differ slightly from our proposed method in that the noise is not annealed and at convergence will be non-zero. 
% They normally differ slightly from our proposed method in that the noise is not annealed and at convergence will be non-zero (corresponding to the Gaussian posterior having finite KL divergence from the prior). 
Additionally, in adaptive weight noise, an extra set of parameters for the variance must be maintained. 

Similarly, the technique of dropout \citep{Srivastava:2014:DSW:2627435.2670313} randomly sets groups of hidden units to zero at train time to improve generalization in a manner similar to ensembling.

An annealed Gaussian gradient noise schedule was used to train the highly non-convex Stochastic Neighbor Embedding model in \cite{hinton2002stochastic}. The gradient noise schedule that we found to be most effective is very similar to the Stochastic Gradient Langevin
Dynamics algorithm of \cite{WellingT11}, who use gradients with added noise to accelerate MCMC inference for logistic regression and independent component analysis models. This use of gradient information in MCMC sampling for machine learning to allow
faster exploration of state space was previously proposed
by~\cite{neal2011mcmc}.

% Stochastic gradient descent~\citep{robbins1951stochastic} is among the
% most efficient optimization algorithms if we consider the
% predictive accuracy obtained per unit of computation. This is because
% the sampling noise induced by the stochastic process can generally
% improve the training of models~\cite{bousquet2008tradeoffs}.

Various optimization techniques have been proposed to improve the
training of neural networks. Most notable is the use of
Momentum~\citep{polyak1964some,sutskever2013importance,kingma2014adam}
or adaptive learning
rates~\citep{duchi2011adaptive,dean2012large,zeiler2012adadelta}. These methods are normally developed to provide good convergence rates for the convex setting, and then heuristically applied to nonconvex problems. 
On the other hand, injecting noise in the gradient is more suitable for nonconvex problems. By adding
even more stochasticity, this technique gives the model
more chances to escape local minima (see a
similar argument in~\cite{bottou1991stochastic}), or to traverse quickly through the ``transient'' plateau phase of early learning (see a similar analysis for momentum in \cite{sutskever2013importance}). 
This is born out empirically in our observation that adding gradient noise can actually result in lower training loss. In this sense, we suspect adding gradient noise is similar to simulated annealing \citep{kirkpatrick1983optimization} which exploits random noise to explore complex optimization landscapes. This can be contrasted with well-known benefits of stochastic gradient descent as a learning algorithm \citep{robbins1951stochastic,bousquet2008tradeoffs}, where both theory and practice have shown that the noise induced by the stochastic process aids generalization by reducing overfitting.

\section{Method}
\label{sec:method}

We consider a simple technique of adding time-dependent Gaussian noise to the gradient $g$ at every training step $t$:
\begin{align*}
g_{t} \leftarrow g_{t} + N(0, \sigma^{2}_{t})
%+ b^{question}
\end{align*}
Our experiments indicate that adding annealed Gaussian noise by decaying the variance works better than using fixed Gaussian noise. We use a schedule inspired from \cite{WellingT11} for most of our experiments and take:
\begin{align}
\label{eq:the_equation}
\sigma^{2}_{t}=\frac{\eta}{(1+t)^\gamma}
%+ b^{question}
\end{align}
with $\eta$ selected from $\{0.01,0.3,1.0\}$ and $\gamma=0.55$.
Higher gradient noise at the beginning of training forces the gradient away from 0 in the early stages.

\section{Experiments}
In the following experiments, we consider a variety of complex neural
network architectures: Deep networks for MNIST digit classification, End-To-End Memory Networks \citep{sukhbaatar2015weakly}
 and Neural Programmer~\citep{neelakantan2016} for question answering, Neural Random Access Machines~\citep{kurach2016} and Neural GPUs~\citep{kaiser2016} for algorithm learning. The models
and results are described as follows.

\subsection{Deep Fully-Connected Networks}

For our first set of experiments, we examine the impact of adding
gradient noise when training a very deep fully-connected network on the
MNIST handwritten digit classification dataset
\citep{lecun1998gradient}.  Our network is deep: it has 20 hidden
layers, with each layer containing 50 hidden units. We use the ReLU
activation function \citep{DBLP:conf/icml/NairH10}.

In this experiment, we add gradient noise sampled from a Gaussian distribution
with mean 0, and decaying variance according to the schedule in Equation \eqref{eq:the_equation}
with $\eta=0.01$. We train with SGD without momentum, using the fixed learning rates of 0.1
and 0.01. Unless otherwise specified, the weights of the network are initialized from a Gaussian with mean zero,
and standard deviation of 0.1, which we call \emph{Simple Init.}

The results of our experiment are in Table~\ref{mnist}. When trained
from Simple Init we can see that adding noise to the gradient helps in
achieving higher average and best accuracy over 20 runs using each learning rate for a total of 40 runs
(Table~\ref{mnist}, Experiment 1). We note that the average is closer
to 50\% because the small learning rate of 0.01 usually gives very
slow convergence. We also try our approach on a more shallow network
of 5 layers, but adding noise does not improve the training in that
case.

Next, we experiment with clipping the gradients with two threshold
values: 100 and 10 (Table~\ref{mnist}, Experiment 2, and 3). Here, we
find training with gradient noise is insensitive to the gradient
clipping values. By tuning the clipping threshold, it is possible to get comparable 
accuracy without noise for this problem.

In our fourth and fifth experiment (Table~\ref{mnist}, Experiment 4, and 5), we use two analytically-derived ReLU 
initialization techniques  (which we term \emph{Good Init}) recently-proposed
by~\cite{DBLP:journals/corr/Sussillo14} and \cite{DBLP:journals/corr/HeZR015}, and find that adding gradient
noise does not help.  Previous work has found that stochastic gradient
descent with carefully tuned initialization, momentum, learning rate,
and learning rate decay can optimize such extremely deep
fully-connected ReLU networks \citep{srivastava2015training}.
It would be harder to find such a robust initialization technique for the more complex
heterogeneous architectures considered in later sections. Accordingly,
we find in later experiments (e.g., Section
\ref{sec:neural_programmer}) that random restarts and the use of a momentum-based optimizer like Adam
are not sufficient to achieve the best results in the absence of added
gradient noise. 

To understand how sensitive the methods are to poor initialization, in addition to the sub-optimal Simple Init, 
we run an experiment where all the weights in the neural network are
initialized at zero. The results (Table~\ref{mnist}, Experiment 5)
show that if we do not add noise to the gradient, the networks fail to
learn. If we add some noise, the networks can learn and reach 94.5\%
accuracy.
\begin{center}
    \begin{table}[h!]
    \begin{tabular}{| l | l | l |}
    \multicolumn{3}{c}{Experiment 1: Simple Init, No Gradient Clipping} \\ \hline
    Setting & Best Test Accuracy & Average Test Accuracy \\ \hline \hline
    No Noise & 89.9\% & 43.1\% \\  \hline
    With Noise & 96.7\% & 52.7\% \\ \hline
    No Noise + Dropout & 11.3\% & 10.8\% \\ \hline 
    \multicolumn{3}{c}{} \\
    \multicolumn{3}{c}{Experiment 2: Simple Init, Gradient Clipping Threshold = 100} \\ \hline
    No Noise & 90.0\% & 46.3\% \\  \hline
    With Noise & 96.7\% & 52.3\% \\ \hline
    \multicolumn{3}{c}{} \\
    \multicolumn{3}{c}{Experiment 3: Simple Init, Gradient Clipping Threshold = 10} \\ \hline
    No Noise & 95.7\% & 51.6\% \\  \hline
    With Noise & 97.0\% & 53.6\% \\ \hline
    \multicolumn{3}{c}{} \\
     \multicolumn{3}{c}{Experiment 4: Good Init~\citep{DBLP:journals/corr/Sussillo14} + Gradient Clipping Threshold = 10} \\ \hline
    No Noise & 97.4\% & 92.1\% \\  \hline
    With Noise & 97.5\% & 92.2\% \\ \hline
    \multicolumn{3}{c}{} \\
    \multicolumn{3}{c}{Experiment 5: Good Init~\citep{DBLP:journals/corr/HeZR015} + Gradient Clipping Threshold = 10} \\ \hline
    No Noise & 97.4\% & 91.7\% \\  \hline
    With Noise & 97.2\% & 91.7\% \\ \hline
    \multicolumn{3}{c}{} \\
    \multicolumn{3}{c}{Experiment 6: Bad Init (Zero Init) + Gradient Clipping Threshold  = 10} \\ \hline
    No Noise & 11.4\% & 10.1\% \\ \hline
    With Noise & 94.5\% & 49.7\% \\  \hline
    \end{tabular}
    \caption{Average and best test accuracy percentages on MNIST over 40 runs. Higher values are better.}
    \label{mnist}
    \end{table}
\end{center}

In summary, these experiments show that if we are
careful with initialization and gradient clipping values, it
is possible to train a very deep fully-connected network without
adding gradient noise. However, if the initialization is poor, optimization
can be difficult, and adding noise to the gradient is a good mechanism
to overcome the optimization difficulty.

The implication of this set of results is that added gradient noise can be an
effective mechanism for training very complex networks. This is
because it is more difficult to initialize the
weights properly for complex networks. In the following, we explore the training of more
complex networks such as End-To-End Memory Networks and Neural Programmer, whose
initialization is less well studied.

\subsection{End-To-End Memory Networks}

We test added gradient noise for training End-To-End Memory
Networks~\citep{sukhbaatar2015weakly}, a new approach for Q\&A using
deep networks.\footnote{Code available at:
  https://github.com/facebook/MemNN } Memory Networks have been
demonstrated to perform well on a relatively challenging toy Q\&A
problem~\citep{weston2015towards}.

In Memory Networks, the model has access to a context, a question, and
is asked to predict an answer. Internally, the model has an attention
mechanism which focuses on the right clue to answer the question. In the original formulation \citep{weston2015towards},
Memory Networks were provided with additional supervision as to what pieces of context were
necessary to answer the question. This was replaced in the End-To-End formulation by a latent
attention mechanism implemented by a softmax over contexts. As this greatly 
complicates the learning problem, the authors implement a two-stage training procedure: 
First train the networks with a linear attention, then use those weights to warmstart the 
model with softmax attention.

In our experiments with Memory Networks, we use our standard noise schedule, using noise sampled from a
Gaussian distribution with mean 0, and decaying variance according to Equation \eqref{eq:the_equation} with $\eta=1.0$.
This noise is added to the gradient after clipping.
We also find for these experiments that a fixed standard deviation
also works, but its value has to be tuned, and works best at 0.001. 
We set the number of
training epochs to 200 because we would like to understand the
behaviors of Memory Networks near convergence. The rest of the
training is identical to the experimental setup proposed by the
original authors. We test this approach with the published two-stage
training approach, and additionally with a one-stage training approach where
we train the networks with softmax attention and without warmstarting. Results
are reported in Table~\ref{memnet}. We find some
fluctuations during each run of the training, but the reported results
reflect the typical gains obtained by adding random noise.

\begin{center}
    \begin{table}[h!]
    \begin{tabular}{| l | l | l |}
    \hline
    Setting            & No Noise & With Noise  \\ \hline \hline
    One-stage training & Training error:   ~~~10.5\%      & Training error:  ~~~~~9.6\%    \\ 
                       & Validation error: 19.5\% & Validation error: 16.6\% \\ \hline
    Two-stage training & Training error:   ~~~~~6.2\%       &  Training error:   ~~~~~5.9\%   \\ 
                       & Validation error: 10.9\% & Validation error: 10.8\% \\ \hline
    \end{tabular}
    \caption{The effects of adding gradient noise to End-To-End Memory
Networks. Lower values are better.}
    \label{memnet}
    \end{table}
\end{center}

We find that warmstarting does indeed help the networks. In
both cases, adding random noise to the gradient also helps the network
both in terms of training errors and validation errors. Added noise,
however, is especially helpful for the training of End-To-End Memory Networks
without the warmstarting stage.

\subsection{Neural Programmer}
\label{sec:neural_programmer}
\begin{figure}[h!]
  \includegraphics[scale=.25]{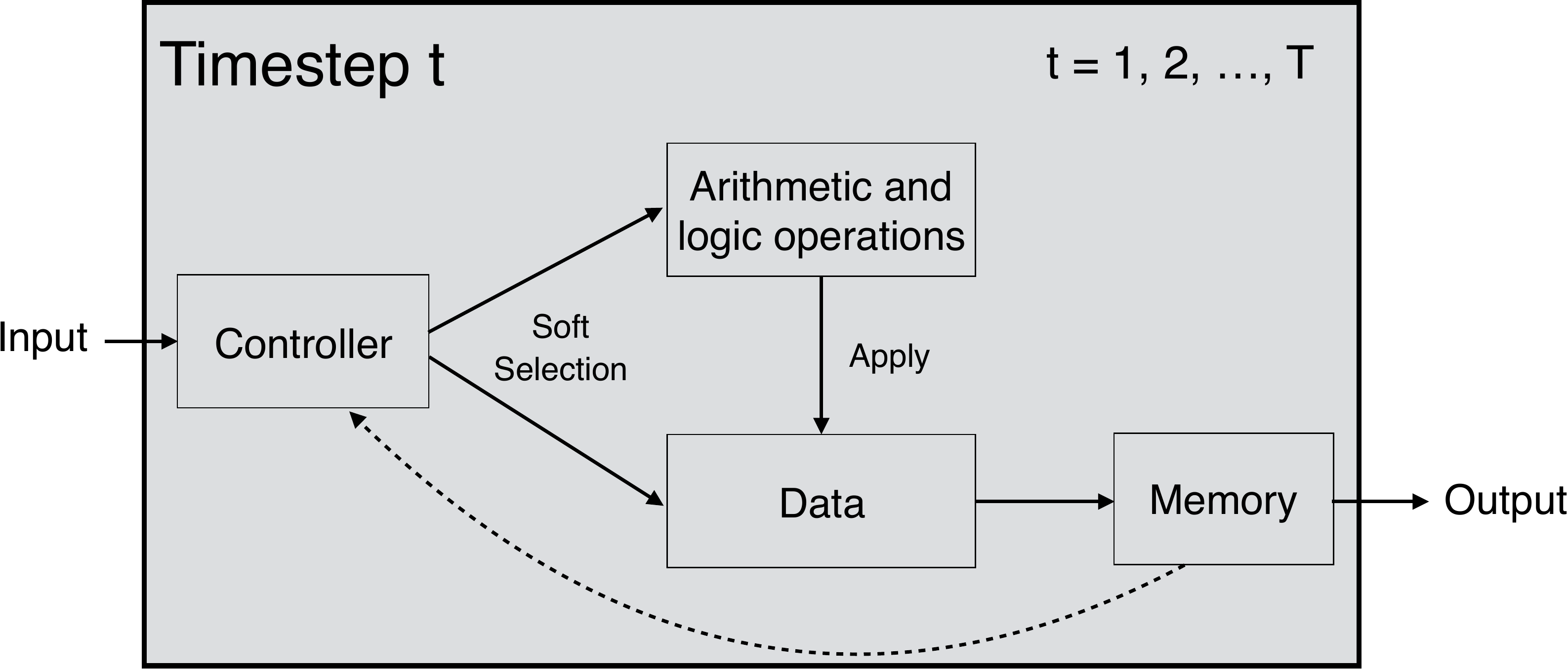}
  \caption{Neural Programmer, a neural network 
    with built-in arithmetic and logic operations. At every time step, the controller selectes an operation and a data segment. Figure reproduced with permission from \cite{neelakantan2016}.}
\label{intro}

\end{figure}

Neural Programmer is a neural network architecture augmented with a
small set of built-in arithmetic and logic operations that learns to
induce latent programs. It is proposed for the task of question answering from tables~\citep{neelakantan2016}. Examples of operations on a table include the sum of a set of numbers, or the list of numbers greater than a particular value. Key
to Neural Programmer is the use of ``soft selection'' to assign a
probability distribution over the list of operations. This
probability distribution weighs the result of each operation, and the
cost function compares this weighted result to the
ground truth. This soft selection, inspired by the soft attention mechanism of~\cite{BahdanauCB14},
allows for full differentiability of the model. Running the model for
several steps of selection allows the model to induce a complex program by chaining
the operations, one after the other. Figure~\ref{intro}
shows the architecture of Neural Programmer at a high level.

In a synthetic table comprehension task, Neural Programmer takes a
question and a table (or database) as input and the goal is to predict the
correct answer. To solve this task, the model has to
induce a program and execute it on the table. A major challenge is
that the supervision signal is in the form of the correct answer and
not the program itself. The model runs for a fixed number of steps, and at
each step selects a data segment and an operation to apply to the selected data segment. Soft selection is performed at training time so that the
model is differentiable, while at test time hard selection is employed. Table \ref{parse} shows examples of programs induced by the model.
\begin{center}
    \begin{table}[h!]
    \begin{tabular}{| c | c | c | c | c |}
    \hline
    {\multirow{2}{*} {Question}} & {\multirow{2}{*} {t}} & Selected & Selected  \\
                                 &                          & Op & Column                                       \\ \hline \hline
    greater 17.27 A and lesser -19.21 D count  & 1 & Greater    & A    \\
    {\it What are the number of elements } & 2  & Lesser    & D  \\
    {\it whose  field in column A is greater than 17.27} & 3 & And   & -  \\
    {\it and  field in Column D is lesser than -19.21.}  & 4 & Count    & -  \\ \hline 
    \end{tabular}
    \caption{Example program induced by the model using $T=4$ time steps. We show the selected columns in cases in which the selected operation acts on a particular column.}
    \label{parse}
    \end{table}
\end{center}
    
Similar to the above experiments with Memory Networks, in our
experiments with Neural Programmer, we add noise sampled from a Gaussian
distribution with mean 0, and decaying variance according to Equation \eqref{eq:the_equation} with $\eta=1.0$ to the gradient after
clipping. The model is optimized with Adam~\citep{kingma2014adam}, which combines momentum and adaptive learning rates. 

For our first experiment, we train Neural Programmer to answer questions involving a single column of numbers. We use $72$ different hyper-parameter configurations with and without adding
annealed random noise to the gradients. We also run each of these
experiments for $3$ different random initializations of the model
parameters and we find that only $1/216$ runs achieve $100\%$ test accuracy
without adding noise while $9/216$ runs achieve $100\%$ accuracy when
random noise is added. The $9$ successful runs consisted of models
initialized with all the three different random seeds, demonstrating robustness to initialization. We find that when using dropout \citep{Srivastava:2014:DSW:2627435.2670313} none of the $216$ runs give 100\% accuracy.

\begin{figure}[h!]
  \centering
  \includegraphics[scale=.35]{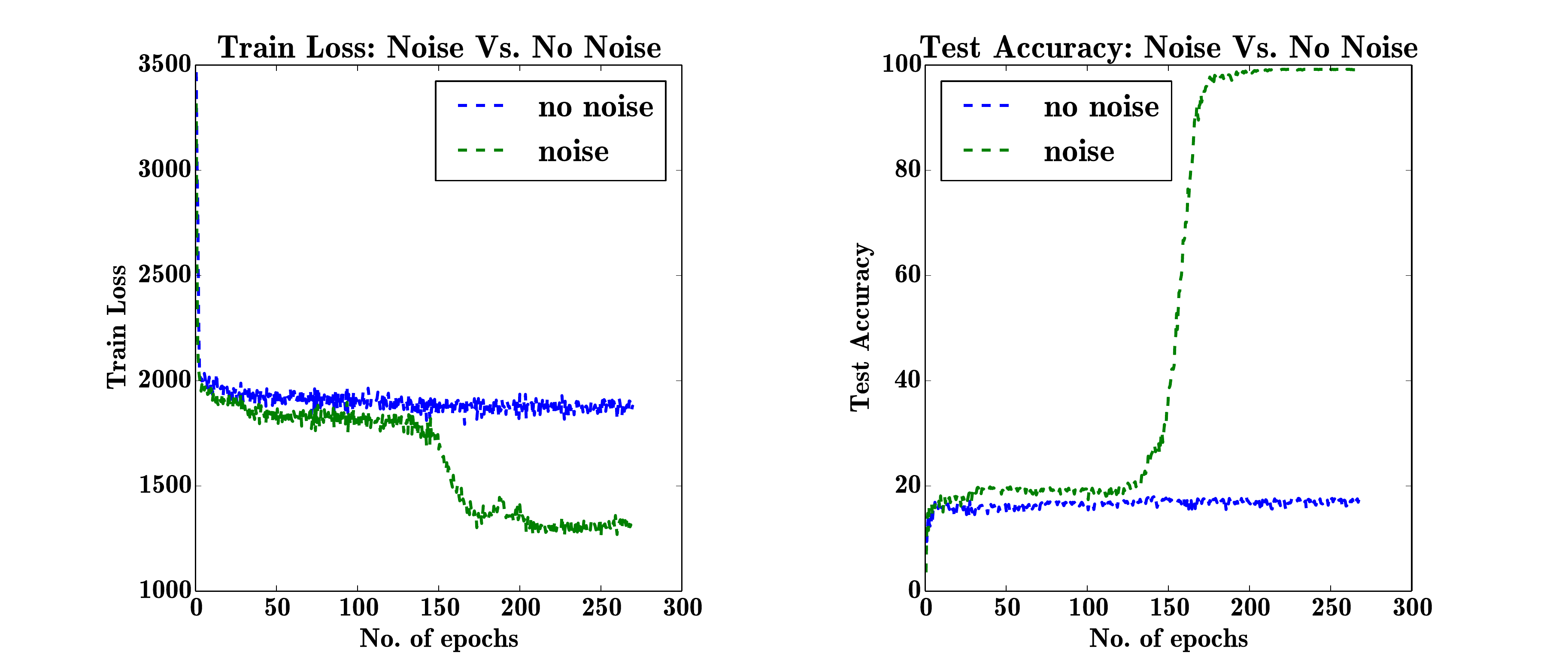}
  \caption{Noise vs. No Noise in our experiment with tables containing 5 columns. The models trained with noise
    generalizes almost always better.}
  \label{noise}
\end{figure}

We consider a more difficult question answering task where tables have
up to five columns containing numbers. We also experiment on a task
containing one column of numbers and another column of text
entries. Table \ref{np} shows the performance of adding noise vs. no
noise on Neural Programmer. 
\begin{center}
    \begin{table}[h!]
    \begin{tabular}{| l | l | l | l| l|}
    \hline
    Setting            & No Noise & With Noise  & Dropout &  Dropout With Noise \\ \hline \hline
    Five columns & 95.3\%     &  98.7\% & 97.4\%  & 99.2\%   \\ \hline
    Text entries  & 97.6\%           &   98.8\% &   99.1\% & 97.3\%   \\ \hline
    \end{tabular}
    \caption{The effects of adding random noise to the gradient on Neural Programmer. Higher values are better. Adding random noise to the gradient always helps the model. When the models are applied to these more complicated tasks than the single column experiment, using dropout and noise together seems to be beneficial in one case while using only one of them achieves the best result in the other case.}
    \label{np}
    \end{table}
\end{center}
Figure \ref{noise} shows an example of the effect of adding random noise to the gradients in our experiment with $5$ columns. The differences between the two models are much more pronounced than in Table \ref{np} because Table \ref{np} shows the results after careful hyperparameter selection.

In all cases, we see that added gradient noise improves performance of Neural Programmer. Its performance when combined with or used instead of dropout is mixed depending on the problem, but the positive results indicate that it is worth attempting on a case-by-case basis. 

% Section about Neural Random-Access Machines
\subsection{Neural Random Access Machines}

We now conduct experiments with Neural Random-Access Machines
(NRAM)~\citep{kurach2016}. NRAM is a model for algorithm learning that can
store data, and explicitly manipulate and dereference pointers. 
NRAM consists of a neural
network controller, memory, registers and a set of built-in
operations. This is similar 
to the Neural Programmer in that it uses a controller network to compose built-in
operations, but both reads and writes to an external memory.
An operation can either read (a subset of) contents from the
memory, write content to the memory or perform an arithmetic operation on either 
input registers or outputs from other operations. 
The controller runs for a fixed number of time steps. At every step, the model 
selects both the operation to be executed and its inputs. 
These selections are made using soft attention \citep{BahdanauCB14} making 
the model end-to-end differentiable. NRAM uses an LSTM \citep{Hochreiter:1997} controller. Figure \ref{fig:nram} gives an overview of the model.

\usetikzlibrary{decorations.pathreplacing,calc,matrix,circuits.logic.US,positioning}

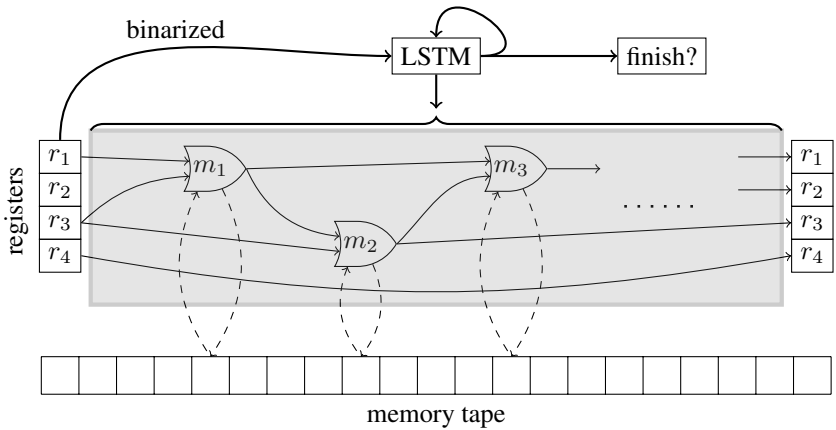
\begin{figure}[h]

\begin{center}
%\framebox[4.0in]{$\;$}
\begin{tikzpicture}[circuit logic US,
                    tiny circuit symbols,
                    every circuit symbol/.style={fill=white,draw, logic gate input sep=2mm}
                    ]
                    
\matrix (R)[matrix of nodes,row sep=-\pgflinewidth,nodes={rectangle,draw}] at (-5,0)
  {
    $r_1$  \\
    $r_2$  \\
    $r_3$  \\
    $r_4$  \\
  };
\node[rotate=90, left=0.3cm of R-1-1] {registers};
  
\node[or gate,draw,point right]  (m1)  at (-3,0.5) {$m_1$};
\draw[->] (R-1-1.east) to[bend right=0] (m1.input 1);
\draw[->] (R-3-1.east) to[bend right=-20] (m1.input 2);

\node[or gate,draw,point right]  (m2)  at (-1,-0.5) {$m_2$};
\draw[->] (m1.output)  to[in=180, bend right] (m2.input 1);
\draw[->] (R-3-1.east) to[bend right=0] (m2.input 2);

\node[or gate,draw,point right]  (m3)  at (1,0.5) {$m_3$};
\draw[->] (m1.output)  to[bend right=0] (m3.input 1);
\draw[->] (m2.output)  to[in=180] (m3.input 2);

\draw[loosely dotted, line width=1pt] (2.5,0) -- (3.5,0);

\matrix (Rp)[matrix of nodes,row sep=-\pgflinewidth,nodes={rectangle,draw}] at (5,0)
  {
    $r_1$  \\
    $r_2$  \\
    $r_3$  \\
    $r_4$  \\
  };

\draw[->] (m3.output)  -- ([xshift=20]m3.output);    
  
\draw[->] ([xshift=-20]Rp-1-1.west)  -- (Rp-1-1);  
\draw[->] ([xshift=-20]Rp-2-1.west)  -- (Rp-2-1);  
\draw[->] (m2.output)  -- (Rp-3-1.west);  
\draw[->,bend right] (R-4-1.east)   to[bend right=10] (Rp-4-1.west);  

\draw [decoration={brace,amplitude=0.5em},decorate,thick]
 (R.north east) --  (Rp.north west);

\node[draw,rectangle] (LSTM) at (0,2) {LSTM};                    
\node[draw,rectangle] (finish) at (3,2) {finish?};
\draw[->,thick] (LSTM.east) to[in=90, out=0, looseness=4] (LSTM.north);
\draw[->,thick] (R-1-1.north) to[in=180, out=90] node[above] {binarized} (LSTM.west); 
\draw[->,thick] (LSTM) to (0,1.3); 
\draw[->,thick] (LSTM) to (finish);

\draw[step=0.5cm,black,xshift=0.25cm] (-5.5,-2.5) grid (5,-2);
\node at (0,-2.8) {memory tape};
\draw[->,dashed] (m1) to[bend left=30] ([yshift=-2.5cm]m1); 
\draw[<-,dashed] (m1) to[bend right=30] ([yshift=-2.5cm]m1); 
\draw[->,dashed] (m2) to[bend left=30] ([yshift=-1.5cm]m2); 
\draw[<-,dashed] (m2) to[bend right=30] ([yshift=-1.5cm]m2); 
\draw[->,dashed] (m3) to[bend left=30] ([yshift=-2.5cm]m3); 
\draw[<-,dashed] (m3) to[bend right=30] ([yshift=-2.5cm]m3); 

\draw [ultra thick, draw=black, fill=gray, opacity=0.2]
       (R.north east) -- (Rp.north west) -- ([yshift=-0.3cm]Rp.south west) -- ([yshift=-0.3cm]R.south east) -- cycle;
 
\end{tikzpicture}
\end{center}
\caption{One timestep of the NRAM architecture with $R=4$ registers and a memory tape. $m_1$, $m_2$ and $m_3$ are example operations built-in to the model. The operations can read and write from memory. At every time step, the LSTM controller softly selects the operation and its inputs. Figure reproduced with permission from \cite{kurach2016}.
\label{fig:nram}
}
\end{figure}

For our experiment, we consider a problem of searching $k$-th element's value
on a linked list. The network is given a pointer to the head of the linked
list, and has to find the value of the $k$-th element. Note that this is highly nontrivial
because pointers and their values are stored at random locations in memory, so the model must
learn to traverse a complex graph for $k$ steps.

% The input to the
% network is represented as a matrix of size $M$ x $M$, where $M$ is the number of
% memory cells.  Each cell is a probability distribution over $1..M$, which can
% be interpreted as a pointer to given memory location. The memory looks as
% follows: $head, k, out, ..., p1, v1, ..., p2, v2, ...$ \\

% The $head$ is a pointer to the first element on the list, $k$ indicates how many hops are needed,
% and the $out$ is a cell where the output should be put. Pairs ($p1, v1$) and ($p2, v2$) represents
% two example elements on the list. For each of them, $p$ points to the next element on the list and $v$ is the
% value. Elements are in random locations in the memory, so that the network needs to follow the pointers to
% find the correct element.  

%More precisly, the input is given as: $head, k, out, ..., p_i, v_i, ..., p_j,
%v_j, ...$ where head is the pointer to the first element on the list, $k$
%indicates how many hops are needed and $out$ is a cell where the output should
%be written. Pairs ($p_i, v_i$) and ($p_j, v_j$) represents two elements from
%the list.

Because of this complexity, training the NRAM architecture can be unstable, especially when the
number of steps and operations is large. We once again experiment with the decaying noise schedule from Equation \eqref{eq:the_equation}, setting $\eta=0.3$. We run a large grid search over the model hyperparameters (detailed in \cite{kurach2016}), and use the top $3$ for our experiments.
For each of these 3 settings, we try 100 different random initializations and look at the percentage of runs that give $100\%$ accuracy  across each one for training both with and without noise.

% We compare the average reproduce ratio for top $3$ hyper-parameter configurations for models trained with and without noise. Each hyper-parameter setting is run with $100$ random restarts. 

As in our experiments with Neural Programmer, we find that gradient clipping is crucial when training with noise. This is likely because the effect of random noise is washed away when gradients become too large.
For models trained with noise we observed much better reproduce rates, 
which are presented in Table~\ref{nram_table}. Although it is possible to train the model to  achieve  $100\%$ accuracy without noise, it is
less robust across multiple random restarts, with over 10x as many initializations leading to a correct answer when using noise.

\begin{center}
    \begin{table}[h!]
    \begin{tabular}{| c | c | c | c | c |}
    \hline
                 & Hyperparameter-1 & Hyperparameter-2 & Hyperparameter-3 & Average\\ \hline \hline
    No Noise & $1$\% & $0$\% & $3$\% & $1.3$\% \\ \hline
    With Noise  & $5$\% & $22$\% & $7$\% & $11.3$\%  \\ \hline
    \end{tabular}
    \caption{Percentage of successful runs on $k$-th element task. Higher values are better. 
    All tests were performed with the same
      set of $100$ random initializations (seeds).}
    \label{nram_table}
    \end{table}
\end{center}

\subsection{Convolutional Gated Recurrent Networks (Neural GPUs)}
Convolutional Gated Recurrent Networks (CGRN) or Neural
GPUs~\citep{kaiser2016} are a recently proposed model that is capable
of learning arbitrary algorithms. CGRNs use a stack of convolution layers, unfolded with tied parameters like a recurrent network. The input data (usually a list of symbols) is first converted to a three dimensional tensor representation containing a sequence of embedded symbols in the first two dimensions, and zeros padding the next dimension. 
Then, multiple layers of modified convolution kernels are applied at each step. The modified kernel is a combination of convolution and Gated Recurrent Units (GRU)~\citep{cho2014learning}. The use of convolution kernels allows computation to be applied in parallel across the input data, while the gating mechanism helps the gradient flow. The additional dimension of the tensor serves as a working memory while the repeated operations are applied at each layer. The output at the final layer is the predicted answer.

The key difference between Neural GPUs and other architectures for algorithmic tasks (e.g., Neural Turing Machines \citep{graves2014neural}) is that instead of 
using sequential data access, convolution kernels are applied in parallel across the input, enabling the use of very deep and wide models. The model is referred to as Neural GPUs because the input data is accessed in parallel. Neural GPUs were shown to outperform previous sequential architectures for algorithm learning on tasks such as binary addition and multiplication, by being able to generalize from much shorter to longer data cases.

In our experiments, we use Neural GPUs for the task of binary multiplication. The input consists two concatenated sequences of binary digits separated by an operator token, and the goal is to multiply the given numbers. During training, the model is trained on 20-digit binary numbers while at test time, the task is to multiply 200-digit numbers.  Once again, we add noise sampled from Gaussian distribution with mean 0, and decaying variance according to the schedule in Equation \eqref{eq:the_equation} with $\eta=1.0$, to the gradient after clipping. The model is optimized using Adam~\citep{kingma2014adam}.

Table \ref{cgrnn} gives the results of a large-scale experiment using Neural GPUs over $7290$ experimental runs. The experiment shows that models trained with added gradient noise are more robust across many random initializations and parameter settings. As you can see, adding gradient noise both allows us to achieve the best performance, with the number of models with $<1\%$ error over twice as large as without noise. But it also helps throughout, improving the robustness of training, with more models training to lower error rates as well.  This experiment shows that the simple technique of added gradient noise is effective even in regimes where we can afford a very large numbers of random restarts.

\begin{center}
    \begin{table}[h!]
    \begin{tabular}{| l | l | l | l | l |}
    \hline
    Setting  & Error $<$ 1\% & Error $<$ 2\% & Error $<$ 3\% & Error $<$ 5\% \\ \hline \hline
    No Noise   & 28 & 90  & 172 & 387 \\ \hline
    With Noise & 58 & 159 & 282 & 570  \\ \hline
    \end{tabular}
    \caption{Number of successful runs on $7290$ random trials. Higher values are better. The models are trained on length $20$ and tested on length $200$.}
    \label{cgrnn}
    \end{table}
\end{center}

\section{Conclusion}

In this paper, we discussed a set of experiments which show the
effectiveness of adding noise to the gradient. We found that adding
noise to the gradient during training helps training and
generalization of complicated neural networks. We suspect that the
effects are pronounced for complex models because they have
many local minima.

We believe that this surprisingly simple yet effective idea,
essentially a single line of code, should be in the toolset of neural
network practitioners when facing issues with training neural
networks. We also believe that this set of empirical results can give
rise to further formal analysis of why adding noise is so effective for very deep neural networks.

\paragraph{Acknowledgements}
%\todo{I added Marcin (in alphabetical order) because of adding NRAM picture. Arvind, Luke - please ACK/modify}\\
We sincerely thank Marcin Andrychowicz, Dmitry Bahdanau, Samy Bengio, Oriol Vinyals for
suggestions and the Google Brain team for help with the project.

\bibliography{noise}
\bibliographystyle{iclr2016_conference}
\end{document}